\documentclass[acmsmall,review=false,screen]{acmart}

\usepackage{bbding}
\usepackage{multirow}
\usepackage{array}
\usepackage{enumitem}
\usepackage{soul}
\usepackage{url}
\usepackage{longtable}
\usepackage{booktabs} 
\usepackage{rotating}
\usepackage{makecell}

\usepackage{graphicx}
\usepackage{tikz}
\usepackage[edges]{forest}
\usetikzlibrary{trees,positioning,shapes,shadows,arrows.meta}

\usepackage{xcolor}

\definecolor{mypp}{RGB}{204,204,250}
\definecolor{myorange}{RGB}{251,235,199}
\definecolor{mygray}{RGB}{208,206,206}
\definecolor{mylightblue}{RGB}{219, 226, 242}
\definecolor{mylightgreen}{RGB}{228, 239, 219}
\definecolor{mylightred}{RGB}{249, 217, 217}
\definecolor{mylightpp}{RGB}{235, 231, 248}

\tikzset{
    basic/.style  = {draw, text width=2.5cm, align=center, font=\sffamily\footnotesize, fill=mylightgreen, rectangle},
    root/.style   = {basic, rounded corners=2pt, text width=1cm, thin, align=center, fill=mypp, text=black},
    level1/.style = {basic, thin, rounded corners=2pt, text width=2cm, align=center, fill=myorange},
    level2/.style = {basic, thin, rounded corners=2pt, align=center, fill=mylightblue, text width=2.65cm, text=black, font=\sffamily\footnotesize},
    edge from parent/.style={draw=black, edge from parent fork right},
}

\usepackage[noend]{algpseudocode}
\usepackage{algorithm}

\AtBeginDocument{%
  \providecommand\BibTeX{{%
    \normalfont B\kern-0.5em{\scshape i\kern-0.25em b}\kern-0.8em\TeX}}}

\newcommand{\revision}[1]{\textcolor{black}{#1}}

\newcommand{\boldparagraph}[1]{\paragraph{\textbf{$\bullet$ #1}}}

\acmJournal{CSUR}
\acmYear{2025} \acmVolume{1} \acmNumber{1} \acmArticle{1} \acmMonth{1} \acmPrice{}\acmDOI{10.1145/3754448}

\begin{document}

\title{Towards Efficient Generative Large Language Model Serving: A Survey from Algorithms to Systems}

\author{Xupeng Miao}
\email{xupeng@purdue.edu}
\affiliation{%
  \institution{Purdue University}
  \country{USA}
}

\author{Gabriele Oliaro}
\email{goliaro@cs.cmu.edu}
\affiliation{%
  \institution{Carnegie Mellon University}
  \country{USA}
}

\author{Zhihao Zhang}
\email{zhihaoz3@cs.cmu.edu}
\affiliation{%
  \institution{Carnegie Mellon University}
  \country{USA}
}

\author{Xinhao Cheng}
\email{xinhaoc@andrew.cmu.edu}
\affiliation{%
  \institution{Carnegie Mellon University}
  \country{USA}
}

\author{Hongyi Jin}
\email{hongyij@cs.cmu.edu}
\affiliation{%
  \institution{Carnegie Mellon University}
  \country{USA}
}

\author{Tianqi Chen}
\email{tqchen@cmu.edu}
\affiliation{%
  \institution{Carnegie Mellon University}
  \country{USA}
}

\author{Zhihao Jia}
\email{zhihao@cmu.edu}
\affiliation{%
  \institution{Carnegie Mellon University}
  \country{USA}
}

\renewcommand{\shortauthors}{Miao, et al.}

\begin{abstract}
  In the rapidly evolving landscape of artificial intelligence (AI), generative large language models (LLMs) stand at the forefront, revolutionizing how we interact with our data.
  However, the computational intensity and memory consumption of deploying these models present substantial challenges in terms of serving efficiency, particularly in scenarios demanding low latency and high throughput.
  This survey addresses the imperative need for efficient LLM serving methodologies from a machine learning system (MLSys) research perspective, standing at the crux of advanced AI innovations and practical system optimizations.
  We provide in-depth analysis, covering a spectrum of solutions, ranging from cutting-edge algorithmic modifications to groundbreaking changes in system designs.
  The survey aims to provide a comprehensive understanding of the current state and future directions in efficient LLM serving, offering valuable insights for researchers and practitioners in overcoming the barriers of effective LLM deployment, thereby reshaping the future of AI.
\end{abstract}

\ccsdesc[500]{Computing methodologies~Machine learning}
\ccsdesc[500]{General and reference~Surveys and overviews}
\ccsdesc[500]{Computing methodologies~Parallel computing methodologies}
\ccsdesc[500]{Computing methodologies~Natural language processing}
\ccsdesc[500]{Hardware~Analysis and design of emerging devices and systems}
\ccsdesc[500]{Computer systems organization~Architectures}
\ccsdesc[500]{Software and its engineering~Operating systems}
\ccsdesc[500]{Hardware~Emerging technologies}

\keywords{large language model, efficiency, algorithm, system, inference, serving}

\received{23 December 2023}
\received[accepted]{16 July 2025}

\maketitle

\section{Introduction}

Generative large language models (LLMs) have become a driving force behind significant advancements in artificial intelligence (AI) and have demonstrated exceptional performance across a wide range of language-related tasks. From machine translation to sentiment analysis, question answering, and text generation, these models have shown their prowess in understanding, generating, and manipulating human languages. 
The advent of Transformer-based architectures, such as GPT-family (Generative Pre-trained Transformer)~\cite{DBLP:journals/corr/abs-2303-08774}, LLaMA-family~\cite{touvron2023llama}, DeepSeek-family~\cite{guo2025deepseek,liu2024deepseek}, and other latest public LLMs (e.g., OPT~\cite{zhang2022opt}, BLOOM~\cite{workshop2022bloom}, Mistral~\cite{jiang2023mistral}, DeciLM~\cite{decilm}, Baichuan~\cite{yang2023baichuan}, GLM~\cite{zeng2022glm}) has played a pivotal role in this paradigm shift, revolutionizing the way natural language processing (NLP) tasks are approached.
Beyond NLP, these models are also transforming a wider range of applications, including automated programming~\cite{chen2021evaluating}, science discovery~\cite{jo2023promise}, personalized digital assistants~\cite{dong2023towards}, creative arts~\cite{ramesh2021zero}, and next-generation computing architecture~\cite{packer2023memgpt}, demonstrating their versatility and profound impact across various industries.

However, the unprecedented success of LLMs has also given rise to several challenges, most notably, their formidable computational requirements during serving. The immense model size and complexity, coupled with the need for extensive computational resources, have impeded their widespread deployment in real-world applications. The resource-intensive nature of these models raises concerns over energy consumption, scalability, and accessibility, hindering their adoption in broader communities without rich compute resources like large companies.

This survey paper aims to address the critical need for efficient LLM serving and presents an exhaustive exploration of the existing multifaceted strategies proposed by the research community to tackle this challenge. We present an in-depth examination of the entire spectrum of solutions, spanning from algorithmic innovations to novel system architectures, all aimed at optimizing the inference process for LLMs.

\subsection{Objectives}

The primary objective of this survey is to provide a comprehensive overview of the latest advancements in LLM serving and inference. We will systematically review and categorize the existing techniques based on their underlying approaches, highlighting their strengths and limitations. The survey will cover a broad range of methodologies, encompassing decoding algorithm, architecture design, model compression, low-bit quantization, parallel computation, memory management, request scheduling, and kernel optimization.

\subsection{Structure}

The paper is structured as follows: Section~\ref{sec:back} introduces the background information about LLM serving. Section~\ref{sec:taxo} includes our taxonomy of existing approaches on efficient LLM serving and revisits these related works from two aspects: algorithmic innovations (\S~\ref{sec:algo}) and system optimizations (\S~\ref{sec:sys}). After that, we list some representative LLM serving frameworks and provide analysis in Section~\ref{sec:comp}. Section~\ref{sec:bench} discusses benchmarks of LLM serving systems. Section~\ref{sec:rela} clarifies the connection between this survey and other related literature. Finally, we propose some promising exploration directions in Section~\ref{sec:future} for improving generative LLM serving efficiency to motivate future research.
\section{Background}
\label{sec:back}
\subsection{Transformer-based LLM}

Transformer-based Large Language Models (LLMs) have marked a significant shift in the field of natural language processing, introducing a new paradigm for understanding and generating human language. Central to this innovation is the Transformer architecture, which is built upon the concept of self-attention mechanisms~\cite{vaswani2017attention}, allowing the model to weigh the importance of different parts of the input data when making predictions.
Mathematically, the self-attention mechanism in Transformers can be described as follows: For an input sequence $X=[x_1, x_2, ..., x_n]$, the Transformer computes a set of queries $Q$, keys $K$ and values $V$ using linear transformations of $X$. The self-attention scores are then computed as:

\begin{equation}
    \text{Attention}(Q, K, V) = \text{softmax}\left(\frac{QK^T}{\sqrt{d_k}}\right)V
\end{equation}

\noindent where $d_k$ is the dimension of the keys. This mechanism allows the model to focus on different parts of the input sequence for each element of the output, capturing complex dependencies regardless of their distance in the input sequence.

Another important structure in Transformers is the Feed-Forward Network (FFN), which is present in each layer of the Transformer and significantly contributes to its computational intensity.
The FFN typically consists of two linear transformations with a non-linear activation function in between, usually represented as:

\begin{equation}
    \text{FFN}(x) = \max(0, xW_1 + b_1)W_2 + b_2
\end{equation}

\noindent Here, $W_1$, $W_2$, $b_1$, and $b_2$ are learnable parameters of the FFN, and the non-linear function $\max(0, \cdot)$ (ReLU, in this case) introduces the necessary non-linearity into the model, allowing it to learn more complex patterns.
The FFN is responsible for a significant portion of the model's parameter count and, consequently, its memory footprint and computational load. In each Transformer layer, after the multi-head attention (MHA) aggregates information from different parts of the input, the FFN processes this aggregated information independently for each position. This parallel processing capability is a key strength of the Transformer, allowing it to handle sequences effectively. However, it also means that the computational load and memory requirements scale with the length of the input sequence and the depth of the network.

The combination of self-attention and FFN in Transformer-based LLMs enables these models to capture a wide range of linguistic contexts and nuances, setting new benchmarks in various NLP tasks. However, the substantial computational requirements for training and inference have become a critical area of research, focusing on optimizing these aspects without significantly compromising performance.
The Transformer model also includes other key components like position encoding, which adds information about the position of each token in the sequence, and the multi-head attention mechanism, which allows the model to focus on different parts of the sequence in different representational spaces.

\subsection{GPUs and Other Accelerators}

The rapid advancement of LLMs owes much to the evolution of GPU architecture and other accelerators, which are integral to enhancing model performance and efficiency. GPUs (Graphics Processing Units) have emerged as a cornerstone in this field, primarily due to their superior parallel processing capabilities. 
Unlike traditional CPUs, which are designed for sequential processing, GPUs consist of thousands of small, efficient cores designed for handling multiple tasks simultaneously. This makes them exceptionally well-suited for the matrix and vector operations that are ubiquitous in deep learning computations, especially for Transformer-based models.

A typical GPU architecture comprises an array of Streaming Multiprocessors (SMs), each containing several cores that share a common instruction unit but can execute independent threads in parallel. 
Additionally, the shared memory (SRAM) within each SM allows for efficient data exchange and synchronization among threads, significantly optimizing the memory access patterns required in LLM computations.
This design is particularly beneficial for the computationally intensive tasks in LLMs, such as the calculations of self-attention and feed-forward networks in Transformers. 
GPUs also come equipped with high-bandwidth memory (HBM), which allows for faster data transfer rates, significantly reducing the bottleneck associated with memory access during large-scale computations.
Moreover, the latest GPU architectures, such as NVIDIA's Ampere and Hopper architectures, continue to offer enhancements and push the boundaries of LLM computation, such as improved memory bandwidth and capacity, higher floating-point operations per second (FLOPS), specialized mixed-precision computing units (i.e., Tensor Core) and more efficient utilization of resources, further accelerating the performance of LLMs. 
Some of them support various precision formats, including FP32 (32-bit floating point), TF32 (TensorFloat-32), FP16 (16-bit floating point), BF16 (Brain Floating Point), and even INT8/INT4, allowing for flexible trade-offs between computational speed and numerical precision, essential in optimizing LLM performance.

Beyond GPUs, a vast array of hardware platforms have been explored for LLM deployment, encompassing CPUs~\cite{shen2023efficient,intelcpu}, mobile and edge devices~\cite{dettmers2023spqr}, ASICs~\cite{peng2023chiplet,zhou2022transpim}, as well as specialized accelerators such as TPUs~\cite{jouppi2023tpu}, FPGAs~\cite{yemme2023scalable}, and other emerging AI chips from various manufacturers (e.g., Apple M2 Ultra~\cite{lai2023relax}, AWS Inferentia~\cite{awsinfer}, SambaNova~\cite{samba}, Cerebras~\cite{dey2023cerebras}, Graphcore IPUs~\cite{graphcore}). 
This survey primarily underscores research anchored in the use of GPUs, and several technical motivations drive this emphasis. 
Due to their architectural innovations and superior computational power, GPUs have dominated the research area of large-scale deep learning in the past few years~\cite{ben2019demystifying}.
Furthermore, the programming languages of GPUs, like NVIDIA's CUDA and AMD's ROCm, facilitate a fine-grained control over thread hierarchies, allowing researchers to exploit the massive parallelism inherent in GPUs.
It attracts numerous developers to build mature software ecosystems on top of these GPUs, fostering a majority of the seminal and advanced LLM research.
While other hardware platforms indeed bring unique strengths to specific contexts, the vast reservoir of research, development, and deployment centered around GPUs makes them an indispensable reference for an in-depth comprehension of LLM inference methodologies.
Considering the hardware similarities, other hardware platforms can also benefit from the design philosophies, insights, and methodologies discussed in this survey.

\subsection{LLM Inference}

LLM inference, particularly in models like GPT (Generative Pre-trained Transformer), often employs an auto-regressive decoding approach. This method is central to how these models generate text, ensuring that each new word or token produced takes into account the entire sequence generated so far. 
Auto-regressive decoding operates under the principle of sequentially predicting the next token in a sequence, given all the previous ones, as shown in Algorithm~\ref{alg:auto}.

\begin{algorithm}
\caption{Auto-Regressive Decoding for LLM Inference}\label{alg:auto}
\begin{algorithmic}[1]
\State Initialize the input sequence \( X_0 \) with a given context or start token
\For{\( t = 1 \) to \( T \)}
    \State Predict the next token \( y_t = \text{argmax}_{y} P(y | X_{t-1}) \)
    \State Update the input sequence \( X_t = X_{t-1} \oplus y_t \)
    \If{$y_t$ is EOS}
        \State \textbf{break}
    \EndIf
\EndFor
\end{algorithmic}
\end{algorithm}

\noindent Here, $P(y | X_{t-1})$ represents the probability of the next token $y$ given the current sequence $X_{t-1}$, and $\oplus$ denotes the concatenation operation. The argmax function is used to select the most probable next token at each step.

This auto-regressive approach is fundamental in LLM inference for generating coherent and contextually appropriate text. It ensures that each token generated is conditioned on a comprehensive understanding of all previously generated content, allowing LLMs to produce highly relevant and fluent text sequences. 
Prior studies have provided in-depth analysis on the algorithmic intensity of Transformer-based LLM inference (e.g., counting the FLOPS, I/O and memory consumption) and extensive empirical results on cost estimation (e.g., modeling the inference latency~\cite{transarith}) according to the auto-regressive decoding algorithm execution.
The optimization of LLM inference is a complex problem as there can be different optimal strategies with different algorithm configurations and system setups.

\subsection{Challenges}

This section describes a variety of challenges for efficient LLM serving.

\boldparagraph{Latency and Response Time}
Efficient large language model inference requires achieving low-latency and fast response times, especially in real-time applications like chatbots, virtual assistants, and interactive systems. Balancing model complexity with inference speed is a critical challenge that necessitates optimizing algorithms and system architectures to minimize response time without compromising accuracy.

\boldparagraph{Memory Footprint and Model Size}
Large language models come with significant memory requirements due to their size and the vast number of parameters they contain. Deploying such models on memory-constrained devices poses a challenge, demanding the development of effective model compression techniques and system optimizations to reduce memory footprint without sacrificing performance.

\boldparagraph{Scalability and Throughput}
Inference systems often face varying levels of request loads in production environments. Ensuring scalability and high throughput to handle multiple simultaneous requests efficiently requires parallel computation, request scheduling, and other system-level optimizations to distribute computational workload effectively across resources.

\boldparagraph{Hardware Compatibility and Acceleration}
Efficiently leveraging hardware resources is crucial for large language model inference. Adapting LLM models to diverse hardware platforms and architectures, including CPUs, GPUs, and specialized accelerators, demands hardware-aware algorithm design and optimization to exploit the full potential of the underlying hardware.

\boldparagraph{Trade-offs between Accuracy and Efficiency}
Optimizing the efficiency of LLM inference may sometimes involve trade-offs with model accuracy. Striking the right balance between model size, computational complexity, and performance is a challenging task that requires careful consideration and evaluation of various algorithmic and system-level techniques.
\section{Taxonomy}
\label{sec:taxo}

Figure~\ref{fig:llm_tax} illustrates our taxonomy of existing efforts on improving the LLM serving efficiency, which can be broadly classified into two categories, including algorithmic innovations and system optimizations. We will discuss each category in details individually.

\begin{figure}[!ht]
    \centering
    \begin{forest}
    for tree={
        grow=east,
        growth parent anchor=west,
        parent anchor=east,
        child anchor=west,
        edge path={\noexpand\path[\forestoption{edge},->, >={latex}] 
            (!u.parent anchor) -- +(5pt,0) |-  (.child anchor) 
            \forestoption{edge label};},
        l sep=10mm, 
        s sep=1mm,  
        fork sep=5mm, 
    }
    [LLM Serving Taxonomy, root
        [System\\Optimizations, level1
            [Kernel Optimizations, level2
                [Automatic Compilation, basic]
                [Variable\\Sequence length, basic]
                [Tailored Attention, basic]
                [Kernel Fusion, basic]
            ]
            [Request Scheduling, level2]
            [Memory Management, level2]
            [Parallel Computation, level2
                [Decentralized Inference, basic]
                [Cloud Scaling, basic]
                [Sequence Parallelism, basic]
                [Model Parallelism, basic]
            ]
            [Low-bit Quantization, level2]
        ]
        [Algorithmic Innovations, level1
            [Model Compression, level2
                [Network Pruning, basic]
                [Knowledge Distillation, basic]
            ]
            [Architecture Design, level2
                [Conditional Computing, basic]
                [Activation Sharing, basic]
                [Recurrent Unit, basic]
                [Attention \\Simplification, basic]
                [Config Downsizing, basic]
            ]
            [Decoding Algorithm, level2
                [Cascade Inference, basic]
                [Early Exiting, basic]
                [Speculative Decoding, basic]
                [Non-autoregressive Decoding, basic]  
            ]
        ]
    ]
    \end{forest}
    \caption{\revision{Taxonomy of LLM Inference Advancements}}
    \label{fig:llm_tax}
\end{figure}

\subsection{Algorithmic Innovation}
\label{sec:algo}

This section presents a comprehensive analysis of the various algorithms and techniques proposed to optimize language model inference efficiency. These works are proposed to address the native performance flaws of large-scale Transformer models through algorithmic advancements.

\subsubsection{Decoding Algorithm} 
\begin{figure*}[t]
    \centering
    \includegraphics[scale=0.32]{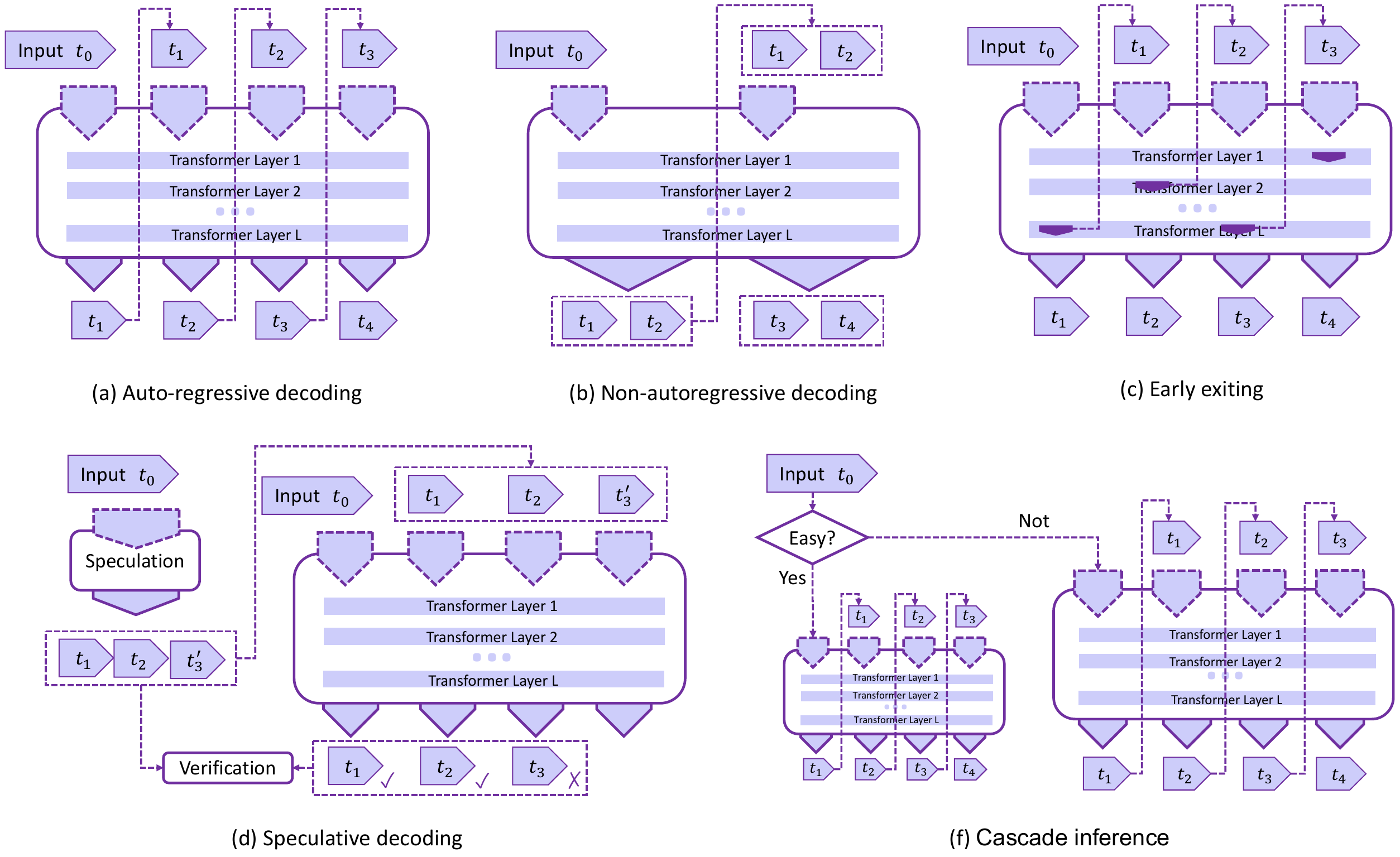}
    \caption{Illustration of different LLM decoding algorithms}
    \label{fig:decoding}
\end{figure*}

In this section, we review novel decoding algorithms as shown in Figure~\ref{fig:decoding} that optimize the inference process of LLMs. These algorithms seek to reduce computational complexity and enhance the overall efficiency of language model inference during generation tasks.

\begin{itemize}
    \item \textbf{Non-autoregressive decoding}. A major limitation of existing LLMs is the default auto-regressive decoding mechanism, which \textit{sequentially} generates output tokens one by one. To address this issue, one representative line of work is to abandon the autoregressive generation paradigm and decode the output tokens \textit{in parallel}.  
    Non-autoregressive decoding~\cite{gu2018non,guo2019non,ghazvininejad2019mask} is first proposed for machine translation acceleration by breaking the word dependencies during decoding and assuming a certain degree of conditional independence.
    To alleviate the translation quality reduction, some follow-up studies like semi-autoregressive decoding~\cite{ghazvininejad2020semi}, further extend these non-autoregressive methods to reach auto-regressive model quality by modeling output dependencies~\cite{gu2021fully,zhan2023depa} or iteratively refining output tokens~\cite{lee2018deterministic}.
    Blockwise parallel decoding~\cite{stern2018blockwise} inserts a single feedforward layer to the base LLM to make predictions for multiple future positions in parallel, then backs off to the longest prefix validated by the base model.
    However, these approaches require to costly reproduce a new LLM with the new dependencies or tune partial layers of the original LLM, which are not always possible.
    Some recent efforts have been dedicated to generate multiple tokens at one decoding step without any training or modification to the model. Parallel decoding~\cite{santilli2023accelerating} reframes the greedy auto-regressive decoding as a system of nonlinear equations solvable in parallel leveraging Jacobi and Gauss-Seidel fixed-point iteration methods for fast inference.
    A thorough survey on non-autoregressive translation~\cite{xiao2023survey} has been proposed to summarize the recent advances in this direction.
    Until now, due to the unawareness of the conditional dependence between output tokens, the output quality of most of non-autoregressive methods has been still less reliable than the auto-regressive method despite an improvement in decoding speed.
    
    \item \textbf{Speculative decoding}. 
    Another line of work addresses the sequential execution limitation by leveraging \textit{speculative execution}~\cite{burton1985speculative} and improving decoding parallelism.
    Each decoding step during the autoregressive LLM inference process can be treated as the execution of a program with conditional branches, such as deciding which token to generate next.
    Speculative decoding~\cite{leviathan2023fast,chen2023accelerating} has been proposed to make decoding predictions of multiple steps first in an efficient manner (e.g., using a smaller draft model with fewer model parameters) and verify these predictions simultaneously with the LLM. 
    However, there are still several practical challenges remaining when applying speculative decoding to LLMs, e.g., how to make decoding predictions light-weight and accurate enough and how to achieve efficient parallel verification using LLMs.
    \revision{As shown in Figure~\ref{fig:specdecoding},} SpecInfer~\cite{miao2023specinfer} first addresses these challenges by introducing a novel tree-based speculative inference and token verification mechanism and proposes a low-latency LLM serving system implementation (\S~\ref{sec:comp}). The main advantage of speculative decoding is that it increases the parallelism without any changes to the outputs.
    Such guarantee comes from the fact that the predicted output is always verified by the original LLM and the fallback mechanism~\cite{kim2023big} takes effect when prediction goes wrong.
    \revision{The tree-based speculative decoding design has been directly adopted by numerous subsequent works, such as Medusa~\cite{cai2023medusa}, EAGLE~\cite{li2024eagle}, and so on~\cite{xu2023llmcad,he2023rest,spector2023accelerating,sun2023spectr,monea2023pass,liu2023online,zhou2023distillspec,zhao2024ouroboros,anknerhydra,oliaro2024suffixdecoding}.
    Some recent efforts~\cite{chen2024sequoia,wang2025opt,li2025adaserve,huang2025specserve} further explore how to adaptively generate better draft token tree structures to improve the speculation efficiency.}
    
    \begin{figure*}[t]
    \centering
    \includegraphics[width=1.0\textwidth]{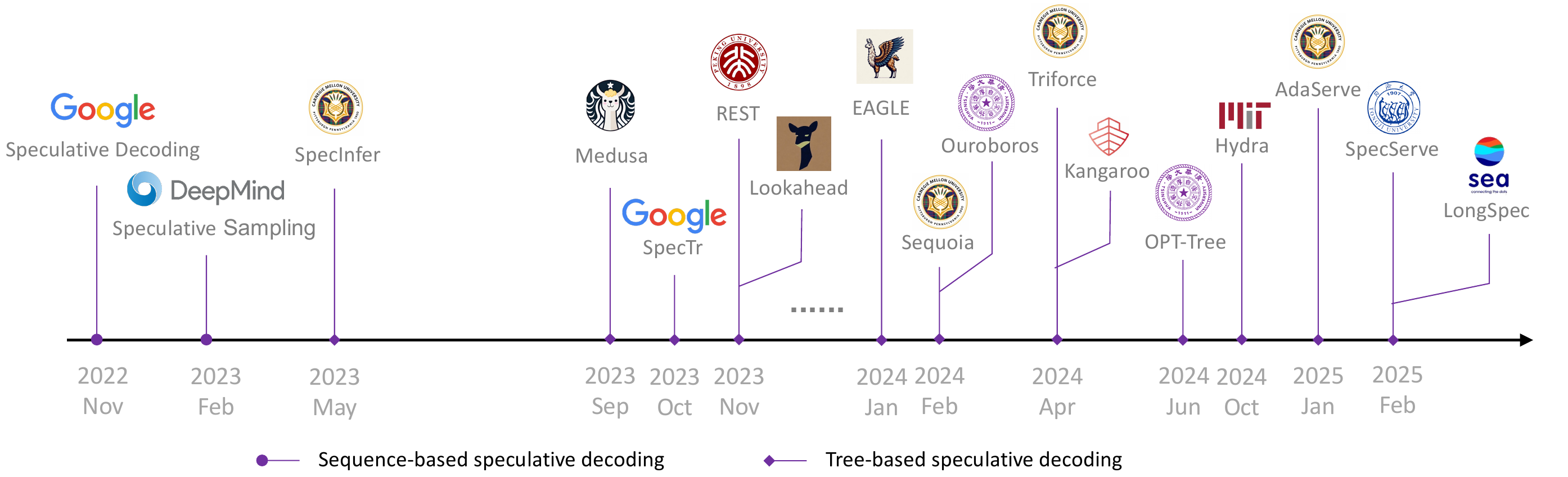}
    \caption{\revision{Illustration of different speculative decoding approaches}}
    \label{fig:specdecoding}
    \end{figure*}

    \item \textbf{Early exiting}. Some other studies attempt to utilize the deep multi-layer architecture of existing LLMs and leverage the \textit{early exiting}~\cite{teerapittayanon2016branchynet} mechanism to accelerate the decoding process.
    The intuition is that the output of early model layers has the potential to infer the target distribution confidently.
    They can emit predictions based on internal classifiers instead of running the whole LLM, and various exit conditions have been explored~\cite{xin2020deebert,liu2020fastbert,zhou2020bert,liao2021global,sun2022simple,he2021magic,kong2022accelerating,ye2021tr,zeng2023learning}.
    They are also called by \textit{adaptive computation}~\cite{schuster2021consistent,del2023skipdecode} since they adjust the amount of computation per request to amortize the total inference cost, i.e., taking less computation for easier inference requests.
    \revision{Kangaroo~\cite{liu2024kangaroo} applies this early exiting strategy to utilize LLM's sub-network and construct a self-drafting model for speculative decoding.}
    Broadly, these approaches are mostly restricted to the insufficient information carried by internal representations and may not faithfully making accurate predictions. 

    \item \textbf{Cascade inference} 
    Driven by the varying complexities of inference requests, cascade inference employs a suite of LLMs of differing scales to minimize response time.
    Instead of directly using a massive model for every query, CascadeBERT~\cite{li2020cascadebert} involves a series of internal classifiers corresponding to different model depths, organizes them in a cascading manner and adaptively selects proper ones based on the instance difficulty.
    Tabi~\cite{wang2023tabi} optimizes for serving discriminative models (i.e., not generative LLMs), but it takes a similar approach to incorporate small models and LLMs to handle queries with different confidence.
    FrugalGPT~\cite{chen2023frugalgpt} leverages a learning-based approach to adaptively assign queries to different LLM APIs, optimizing both cost and performance.
    A concurrent work~\cite{zhu2023optimal} jointly optimizes model multiplexing and query caching and also analyzes the optimality of minimizing inference cost.
    Mixture-of-thought~\cite{yue2023large} extends the cascade idea to LLM reasoning tasks for cost-saving, which samples answers from both Chain-of-Thought~\cite{wei2022chain} and Program-of-Thought~\cite{chen2022program} prompts.
    Overall, cascade inference is a promising direction for enhanced inference efficiency, but it is still challenging to design an accurate dispatching mechanism to avoid compromising model quality.
    
\end{itemize}

\subsubsection{Architecture Design} This subsection explores innovative architecture designs tailored for large language models. Researchers have proposed novel model architectures~\cite{he2023simplifying} beyond the original Transformer that strike a balance between model size, performance, and efficiency, opening new avenues for faster and resource-efficient inference.

\begin{itemize}
    \item \textbf{Configuration downsizing}: 
    To reduce the computation cost of LLM inference, a straightforward approach is to downsize the model configurations, such as using shallow encoders~\cite{goyal2020power,modarressi2022adapler} or decoders~\cite{kasai2020deep}, weight sharing, and vocabulary shrinking~\cite{shi2017speeding}. However, reducing the number of model parameters also affects the downstream tasks' performance.
    \item \textbf{Attention simplification}: One prominent challenge associated with self-attention calculations is the computational complexity $\mathcal{O}(L^2)$, which scales quadratically with the input sequence length $L$.
    Numerous Transformer variants~\cite{DBLP:journals/csur/TayDBM23} have been proposed to simplify the standard attention into more efficient alternatives for very long sequence tasks, such as sparsification~\cite{zaheer2020big}, kernelization~\cite{katharopoulos2020transformers}, and factorization~\cite{wang2020linformer}.
    \revision{Recently, there is a trend of borrowing the ideas from prior attention simplification approaches, generalizing and combining them to shorten the context or reduce the size of KV cache, as well as the attention complexity, with slightly decoding quality degradation (e.g., sliding window attention~\cite{jiang2023mistral,zhang2023efficient}, hash-based attention~\cite{pagliardini2023faster,chen2024magicpig}, dilated attention~\cite{ding2023longnet}, product quantization~\cite{zhang2024pqcache}).}
    One category of these approaches is \textit{context compression} by compressing the context into fewer \textit{soft} tokens (e.g., replacing with summary tokens~\cite{chevalier2023adapting} or landmark tokens~\cite{mohtashami2023landmark}, leveraging additional autoencoder schemes~\cite{liu2023cachegen,ge2023context}) or directly dropping or rephrasing unimportant context tokens based on different importance guidance~\cite{li2023compressing,jiang2023llmlingua,mu2023learning,fei2023extending} (or called \textit{semantic compression}).
    For example, adaptively sparse attention~\cite{anagnostidis2023dynamic} takes a learning-based approach to eliminate uninformative context tokens dynamically for each token. Scissorhands~\cite{liu2023scissorhands} and H$_2$O~\cite{zhang2023h} select a few important tokens that might have a substantial influence for future decoding process and save their KV cache.
    StreamingLLM~\cite{xiao2023efficient} values the initial tokens and maintains them with the sliding window, which is also similar to prior  work~\cite{beltagy2020longformer}.
    \revision{TriForce~\cite{suntriforce} extends this idea to reduce the drafting latency and proposes a hierarchical speculative decoding algorithm.}
    FastGen~\cite{ge2023model} allows different attention heads to employ different emphasizing patterns adaptively.
    Table~\ref{tab:atten} illustrates the sparse attention patterns of four representative categories of approaches and their applications.
    However, due to the incomplete context, these approaches may face inevitable information loss in real workloads with more complex attention distributions.

\begin{table}[t]
\centering
\caption{Comparisons of attention simplification methods in prior efficient Transformers and recent LLMs.}
\label{tab:atten}
\resizebox{\textwidth}{!}{
\begin{tabular}{@{}l|c|c|c|c@{}} 
\toprule
Attention Type & Selective & Sliding + Dilated & Global token & Hash-based \\ 
\midrule
Sparse Pattern & \includegraphics[width=0.15\textwidth]{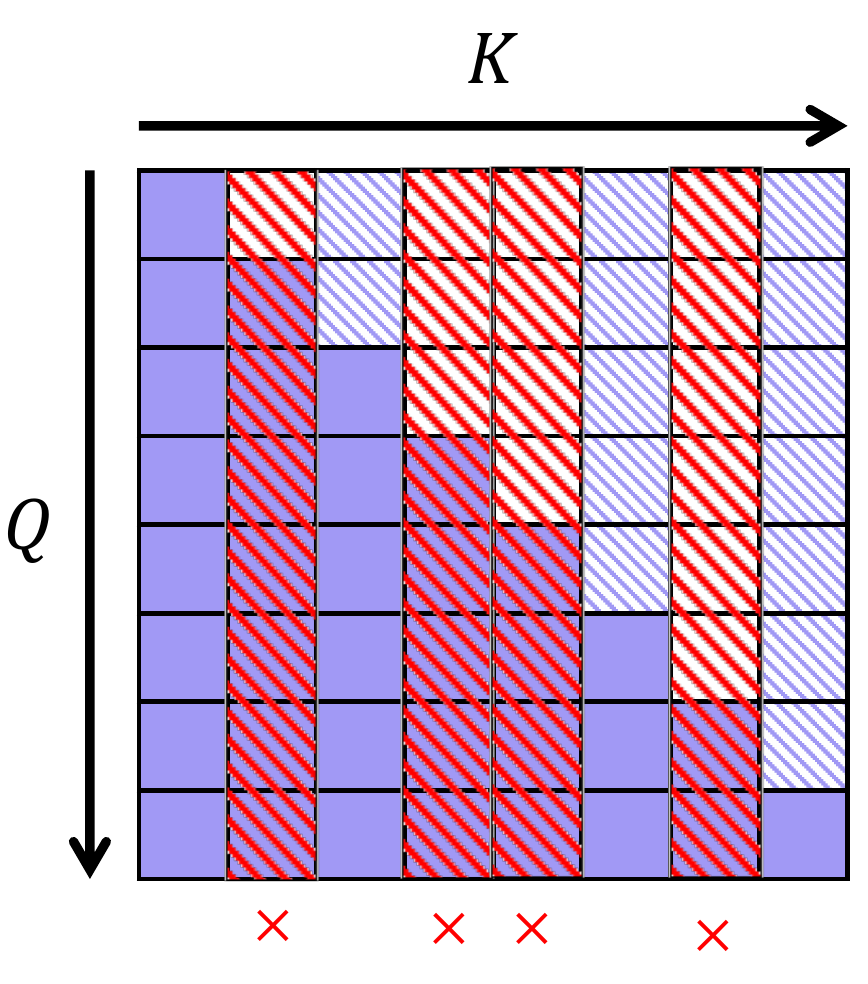} & \includegraphics[width=0.15\textwidth]{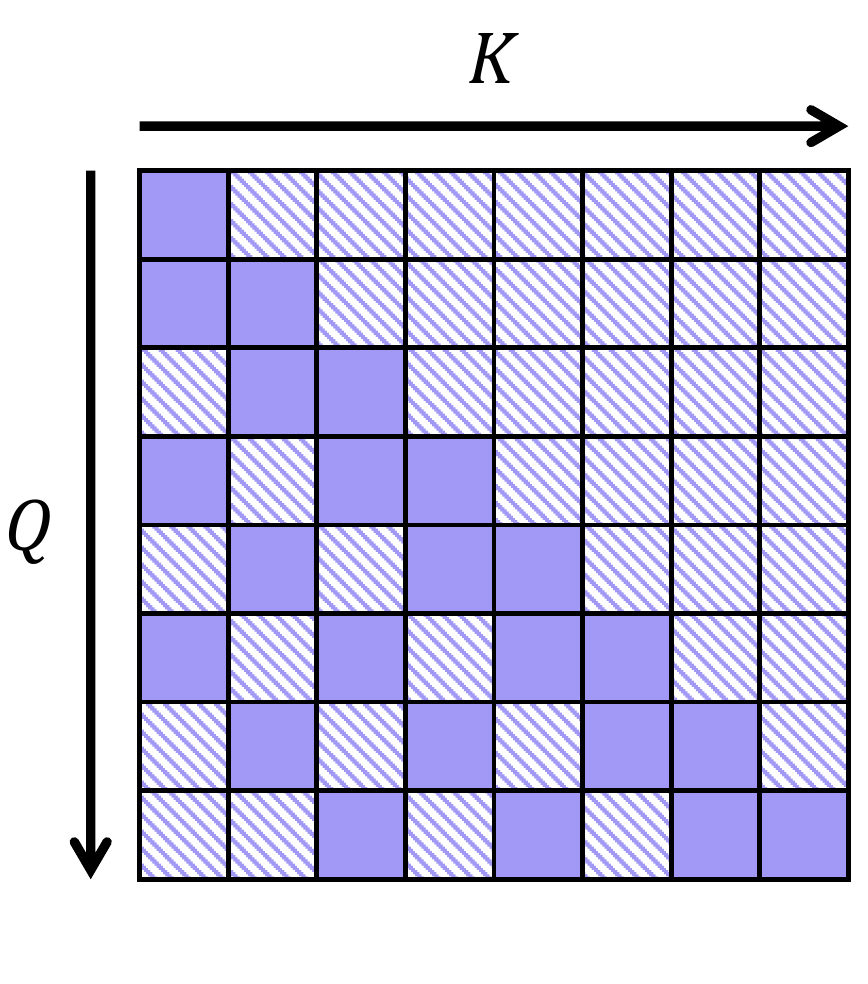} & 
\includegraphics[width=0.15\textwidth]{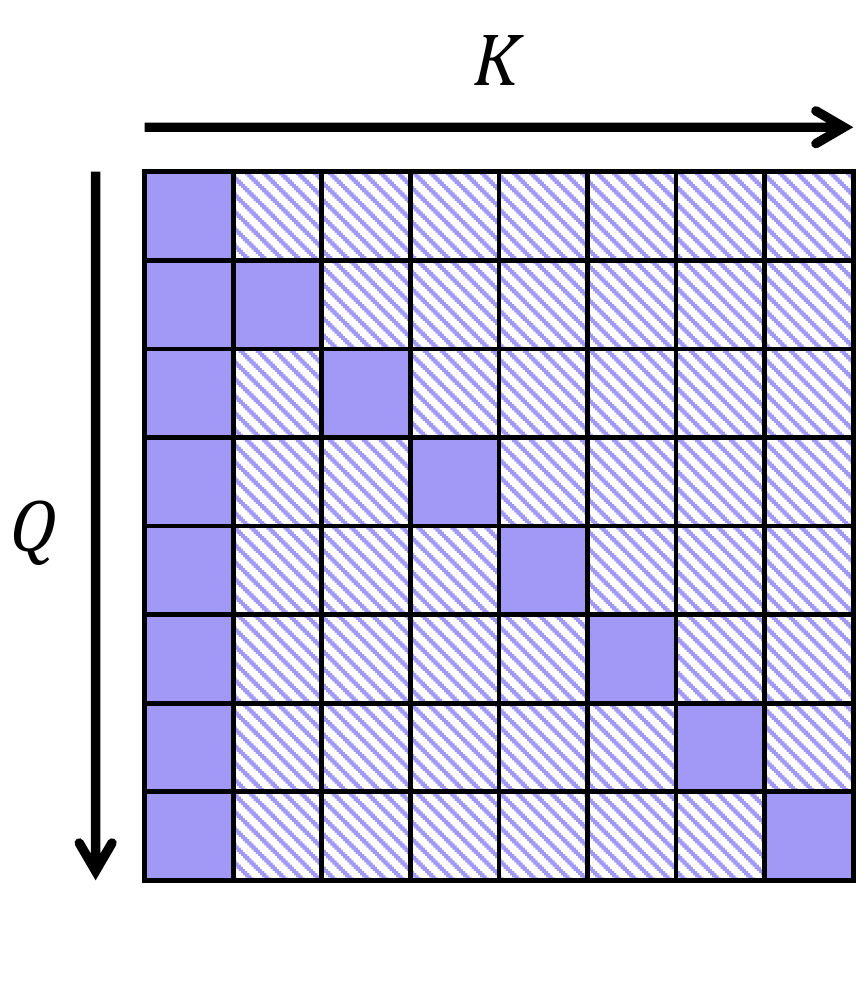} & \includegraphics[width=0.15\textwidth]{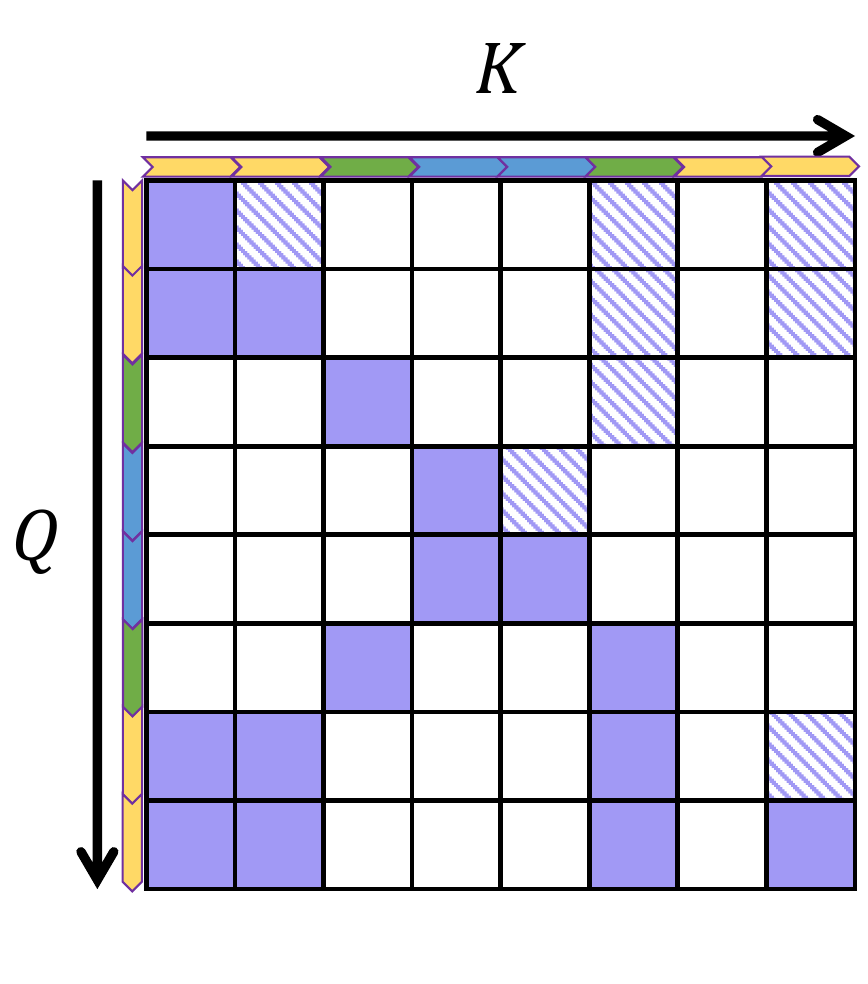} \\ 
\hline
\multirow{2}{*}{References} & Top-$k$~\cite{gupta2021memory}, Sorting~\cite{tay2020sparse}, & Sparse Transformer~\cite{child2019generating}, & Star Transformer~\cite{guo2019star}, & Reformer~\cite{kitaev2019reformer}, Routing\\
& Adaptive~\cite{correia2019adaptively}, Informer~\cite{zhou2021informer} & LongFormer~\cite{beltagy2020longformer} & GMAT~\cite{gupta2020gmat} &  Transformer~\cite{roy2021efficient}\\ \hline
LLM & Scissorhands~\cite{liu2023scissorhands}, H$_{2}$O~\cite{zhang2023h} & Mistral-7B~\cite{jiang2023mistral}, & StreamingLLM~\cite{xiao2023efficient}, & Sparse hash \\ 
Applications & \cite{li2023compressing,jiang2023llmlingua,mu2023learning,anagnostidis2023dynamic,zhang2023h} & \cite{xiao2023efficient}, LongNet~\cite{ding2023longnet} & Summary~\cite{chevalier2023adapting}, Landmark~\cite{mohtashami2023landmark} & attention\cite{pagliardini2023faster} \\
\bottomrule
\end{tabular}
}
\end{table}
    
    \item \textbf{Activation sharing}: Another direction is sharing the intermediate activations to improve the attention calculation efficiency.
    Attention sharing approaches~\cite{DBLP:conf/ijcai/XiaoLZ0L19,wu2022speeding,li2021efficient} observe the similarity among different layers' attention matrix distribution and reuse these attention matrices to reduce the computation costs.
    Multi-query attention (MQA)~\cite{shazeer2019fast} makes different heads share a single set of keys and values to reduce the memory bandwidth requirements in the incremental inference.
    Group-query attention (GQA)~\cite{ainslie2023gqa} relaxes the single set of keys and values restriction to multiple sets and each set is coupled with a group of queries.
    \revision{Multi-Head Latent Attention (MLA)~\cite{liu2024deepseek} jointly compresses keys and values in a low-rank latent vector, leading to significant memory reduction.}
    They have been successfully adopted by several recent public LLMs and shown their superior performance, including, MQA-based models such as Falcon~\cite{zxhang2023falcon}, PaLM~\cite{chowdhery2022palm}, ChatGLM2-6B~\cite{chatglm2}, GQA-based models like LLaMA-2~\cite{touvron2023llama} and Mistral-7B~\cite{jiang2023mistral}, and MLA-based models such as DeepSeek V2, V3, and R1~\cite{guo2025deepseek}.
    
    \item \textbf{Conditional computing}: 
    The sparsely-activated Mixture of Experts (MoE)~\cite{shazeer2017outrageously,csordas2023switchhead} paradigm partitions a model's capacity across various ``experts'', which are smaller neural networks, each specializing in different subsets of the data.
    It allows the system to only invoke the necessary experts for a given input based on certain routing mechanisms~\cite{fedus2022switch,lepikhin2020gshard,nie2021evomoe,roller2021hash,zhou2022mixture,santos2023memory}, rather than computing over the entire massive model, yielding computational and memory efficiency~\cite{du2022glam}.
    For example, TaskMoE~\cite{kudugunta2021beyond} illustrates that task-level routing enables model increase capacity compared with token-level counterpart, while improving the inference throughput.
    As LLMs continue to grow, the MoE architecture stands out as a promising avenue to ensure both scalability and efficiency for future LLMs.
    In the meanwhile, the dynamic nature of MoEs also demands special system optimization from both distributed communication~\cite{he2022fastermoe,nie2023flexmoe,rajbhandari2022deepspeed,hwang2023tutel,li2023accelerating,huang2023towards,deepep2025} and GPU kernel implementation~\cite{gale2023megablocks,zheng2023pit} to facilitate MoE inference efficiency.
    
    \item \textbf{Recurrent unit}: Although recurrent neural networks (RNN) (e.g., LSTM~\cite{sak2014long}) tend to struggle with capturing long-term dependencies in sequences~\cite{khandelwal2018sharp}, there are still several approaches using recurrent units to replace Transformer modules and achieve linear computational and memory complexity during inference, such as RWKV~\cite{peng2023rwkv} and RetNet~\cite{sun2023retentive}.
    Specifically, unlike prior approaches, these recent explorations are mostly built on the linear attention (i.e., Linear Transformer~\cite{katharopoulos2020transformers}, Attention Free Transformer~\cite{zhai2021attention}) representation. 
    After the reformation, they overcome the $\mathcal{O}(L^2)$ bottleneck of attention by modeling interactions between tokens with linear recurrence units (e.g., state space models~\cite{gu2021efficiently,mehta2022long,fu2022hungry,gu2023mamba}, LRU~\cite{orvieto2023resurrecting}), which are easier to maintain parallelizable training property.
    Their design is also composed of various position encoding modules~\cite{su2021roformer}, exponential decay mechanisms~\cite{oliva2017statistical} and a stack of token-wise non-linear MLPs~\cite{yu2022metaformer,tolstikhin2021mlp} or GLUs~\cite{dauphin2017language} to improve the model representation capability.
    Recently, they have shown promising results on both model performance and computation efficiency. However, whether recurrent units can successfully replace Transformers for LLMs still remains an open problem (i.e., especially for long sequences).

\end{itemize}

\subsubsection{Model Compression} Here, we delve into techniques for model compression, which aim to reduce the memory footprint and computational requirements of LLMs by creating more efficient and compact models without significant loss in performance.

\begin{itemize}
    \item \textbf{Knowledge Distillation}: One line of work is knowledge distillation, which trains a small student model with the supervision from a large teacher model. Most previous approaches in this direction are exploring white-box distillation~\cite{sanh2019distilbert,sun2019patient,jiao2020tinybert,wang2020minilm,gu2023knowledge}, which require accessing the entire teacher model parameters. 
    Due to the arising of API-based LLM services (e.g., ChatGPT), several black-box distilled models attract lots of attention, such as Alpaca~\cite{taori2023stanford}, Vicuna~\cite{vicuna2023}, WizardLM~\cite{xu2023wizardlm} and so on~\cite{peng2023instruction,zhu2023minigpt}. These models usually have fewer model parameters but have shown promising performance on various downstream tasks compared with the original LLMs (e.g., GPT-4~\cite{DBLP:journals/corr/abs-2303-08774}).
    
    \item \textbf{Network pruning}: Network pruning methods~\cite{sanh2020movement,michel2019sixteen,sanh2020movement} have been extensively studied in the past few years but not all of them can be directly applied to LLMs.
    It is imperative to take into account the potentially exorbitant computational costs associated with retraining, as well as assess whether the pruning yields discernible gains in inference efficiency based on the underlying system's implementation.
    Some recent approaches~\cite{ma2023llm,santacroce2023matters,fan2019reducing,kurtic2023ziplm} apply structural pruning methods on LLMs, which removes entire structured LLM components, facilitating efficient GPU speedups.
    For example, Deja Vu~\cite{liu2023deja} cuts off specific attention heads and MLP parameters guided by the contextual sparsity hypothesis without modifying pre-trained models.
    There are also some recent advancements in unstructured methods~\cite{frantar2023sparsegpt,xu2023compress,sun2023simple,valicenti2023mini,belcak2023exponentially}, which usually achieve 50-60\% sparsity for LLM compression.
    It is noteworthy that they can further generalize to semi-structured N:M sparsity (i.e., 2:4 and 4:8)~\cite{mishra2021accelerating}, leading to significant inference speedup with NVIDIA sparse tensor cores' acceleration. 
    LoSparse~\cite{li2023losparse} and DSFormer~\cite{chand2023dsformer} approximate model weights with a small dense and a sparse semi-structured matrix using low-rank factorization.
    Flash-LLM~\cite{xia2023flash} relaxes this requirement by providing a memory-efficient SpMM implementation for unstructured pruning using tensor cores.
    PowerInfer~\cite{song2023powerinfer} assumes skew access of these sparsely-activated neurons and proposes a GPU-CPU hybrid inference engine, making GPU and CPU handle different neurons.
    
\end{itemize}

\subsection{System Optimization}
\label{sec:sys}

This section investigates LLM inference system optimization techniques to accelerate LLM inference without modifying the LLM computation semantics. The goal of this line of work is to improve the system efficiency by refining the underlying systems and frameworks used for large language model inference.

\subsubsection{Low-bit Quantization} This section explores state-of-the-art low-bit quantization techniques that enable efficient representation of model weights and activations. By using fewer bits (i.e., less than 32) to represent numerical values, these methods significantly reduce memory consumption and accelerate inference on hardware platforms.
One line of approach is to quantize LLM, and these quantization methods can be briefly categorized into two directions: Quantization-Aware Training (QAT) and Post-Training Quantization (PTQ)~\cite{yao2023comprehensive}.
PTQ reduces the computational precision of model weights~\cite{frantar2022gptq,frantar2022optq,dettmers2022llmint8,lin2023awq,dettmers2023spqr,isik2023gpt} and even activations~\cite{yao2022zeroquant,xiao2022smoothquant,yuan2023rptq} into either INT8 or INT4 by using custom CUDA kernels~\cite{park2022nuqmm,li2023speed} or compilations~\cite{zhao2023atom} for efficiency benefits, such as W8A16 (i.e., INT8 weight-only quantization and FP16 or BF16 activations), W4A16 in GPTQ~\cite{frantar2022gptq}, W8A8 in SmoothQuant~\cite{xiao2022smoothquant} and W4A4~\cite{wu2023understanding}.
The evolution of hardware also meets these requirements. One supporting evidence is that NVIDIA's recent architectures like Turing and Ampere have included INT8 and INT4 tensor cores, and the latest Hopper architecture has disabled INT4 support but introduced FP8 tensor cores for better numerical precision (e.g., H100 GPU can reach 60$\times$ TFLOPS for FP8 as opposed to FP32).
Existing approaches usually adopt various quantization functions, including uniform methods (i.e., Round-to-Nearest) and non-uniform methods~\cite{kim2023squeezellm}.
To relieve the performance loss from low-precision, QAT integrates quantization during model training~\cite{liu2023llm,dettmers2023qlora}.
It is worth noting that due to challenges in the underlying system implementation, low-precision quantization methods may potentially result in slower inference speeds compared to conventional precision levels such as FP16~\cite{dettmers2022llmint8}. While low-precision methods significantly reduce the resource requirements for model deployment, there is also research indicating that quantization methods can have a notable impact on the model's inference performance due to the presence of scaling laws~\cite{dettmers2022case}.
In addition, quantization has also been applied to context compression (e.g., CacheGen~\cite{liu2023cachegen}) and memory-efficient fine-tuning (e.g., QLoRA~\cite{dettmers2023qlora}, PEQA~\cite{kim2023memory}), resulting in lower memory consumption for LLM inference.

\subsubsection{Parallel Computation} 
This section examines parallel computation strategies tailored for large language models. Leveraging parallel processing capabilities of modern hardware architectures, these methods distribute computation across multiple cores or devices, leading to substantial speedup during inference.

\begin{itemize}
    \item \textbf{Model parallelism}: Most model parallelism approaches are first proposed for distributed training of large-scale DNNs, especially for Transformer-based models.
    For example, tensor model parallelism~\cite{shoeybi2019megatron} (TP) splits the model layers (e.g., attention, FFN) into multiple pieces from internal dimensions (e.g., head, hidden) and deploys each on a separate device (e.g., GPU).
    It can significantly reduce inference latency through parallel computing, which is widely used for multiple GPUs within the same machine, especially for scenarios with high-speed NVLink connections.
    PaLM inference~\cite{pope2023efficiently} extends TP on large-scale Transformer inference by involving 2D tensor parallelism~\cite{van1997summa} and claims lower theoretical communication complexity for large clusters (more than 256 devices). For MQA with only one head for keys and values, it further involves data parallelism to the hybrid tensor partition strategy.
    Pipeline model parallelism~\cite{narayanan2021memory,jeon2025graphpipe} (PP) arranges the model layers in a sequence across multiple devices. Each device is responsible for a pipeline stage that consists of multiple consecutive model layers.
    While PP can significantly increase the number of inputs processed per unit of time (throughput), it doesn't inherently decrease the time taken to process a single input from beginning to the end (latency) like TP.
    Different parallelism techniques introduce varying degrees of communication overhead and computational latency~\cite{isaev2023calculon}.
    To achieve optimal performance and resource utilization, automatic parallelism has been widely studied by prior approaches for distributed training (e.g., Alpa~\cite{zheng2022alpa}, FlexFlow~\cite{jia2019beyond,unger2022unity}, Galvatron~\cite{miao2023galvatron}).
    By replacing their cost model to fit the predictable runtime of auto-regressive inference of Transformer models like~\cite{narayanan2023cheaply}, it's easy to apply previous automatic searching algorithms (e.g., dynamic programming, integer linear programming) to LLM serving (e.g., AlpaServe~\cite{li2023alpaserve}, FlexFlow-Serve~\cite{flexflowserve}, SpotServe~\cite{asplos24spotserve}) and determine the most efficient parallelism strategy without manual intervention.

    \item \revision{\textbf{Sequence parallelism}:
    As LLMs increasingly handle tasks requiring extensive context, such as document analysis or conversational agents, processing lengthy sequences efficiently becomes critical.
    Sequence parallelism (SP) distributes the computational and storage load by splitting the processing of long sequences across multiple GPUs along the sequence length dimension.
    Each GPU processes a portion of the sequence, computes local attention outputs, and then communicates the necessary intermediate results (using techniques such as ring-style~\cite{liu2023ring,brandon2023striped} or all-gather~\cite{korthikanti2023reducing,jacobs2023deepspeed} operations) to ensure that the full sequence context is incorporated into the final output.
    Meta further explores the implementation details of context parallelism~\cite{yang2024context} and proposes two ring attention variants for lower latency under varying context lengths and KV cache hit rates.
    To handling the growing context window, LoongServe~\cite{wu2024loongserve} introduces elastic sequence parallelism, automatically  scaling-up and scaling-down according to the context length changes.}

    \item \revision{\textbf{Cloud scaling}: Recent advancements in cloud scaling have significantly enhanced the scalability and cost-effectiveness  of serving LLMs. One notable approach involves utilizing preemptible or spot instances --- cost-effective cloud resources that can be reclaimed by providers with minimal notice.
    SpotServe~\cite{asplos24spotserve} is a distributed LLM serving system designed to operate on preemptible instances, dynamically adjusting parallelization configurations, and migrating GPU context to maintain reliable performance despite potential instance preemptions.
    Similarly, SkyServe~\cite{mao2024skyserve} leverages a combination of spot instances across various regions and clouds to enhance availability and mitigate correlated preemptions. 
    Another innovative direction is the application of serverless computing to LLM serving.
    ServerlessLLM~\cite{fu2024serverlessllm} the cold start challenge by implementing a multitier checkpoint loading system, leveraging underutilized GPU memory and storage to reduce the startup latency.
    Collectively, these advancements demonstrate the potential of integrating cloud computing paradigms to optimize the scalability, reliability, and cost-efficiency of LLM serving.}
    
    \item \textbf{Decentralized inference}: 
    \revision{Serving LLMs in decentralized environments leverages geo-distributed or even heterogeneous resources to perform inference tasks. This approach utilizes idle GPU resources across a network of devices, allowing for efficient processing of LLM workloads without relying on an expensive centralized infrastructure. Decentralized LLM serving addresses challenges such as high costs and limited availability of GPU resources, making it a viable alternative to traditional centralized systems.}
    Inspired by crowdsourced computing, Petals~\cite{borzunov2022petals} serves a BLOOM-176B model using collaborated commodity GPUs over the Internet.
    \revision{Helix~\cite{mei2025helix} optimizes the model placement and request scheduling under geo-distributed heterogeneous environments with a max-flow-based approach for more affordable LLM serving.}
    Decentralized inference opens up a new direction on unlocking the overlooked consumer-level GPUs for running LLMs, but also suffers from several practical challenges, such as device heterogeneity~\cite{jiang2023hexgen}, limited computational and memory capacity~\cite{jiang2025demystifying}, low-bandwidth network~\cite{borzunov2023distributed}, fault tolerance and privacy protection~\cite{tang2023fusionai}.
    
\end{itemize}

\subsubsection{Memory Management}
\label{sec:mem}
Efficient memory management remains at the forefront of challenges in LLM serving, especially given the inherent memory-intensive nature of transformer architectures.
With the growing need for long-sequence inference, the memory footprint of the KV cache stands out as a prime optimization target compared with model weights and the necessary workspace for other activations.
As the KV cache memory grows and shrinks dynamically and unpredictably during incremental decoding, the naive approach (e.g., FasterTransformer) pre-allocates a contiguous piece of memory with a maximum sequence length assumption.
It wastes memory severely for 1) input batches with varied request lengths and 2) complex decoding scenarios generating multiple output sequences in parallel (e.g., beam search, parallel decoding).
vLLM~\cite{kwon2023efficient} proposes \textit{paged attention} that partitions the KV cache into non-contiguous memory blocks and significantly improves the batch size as well as throughput.
\revision{vAttention~\cite{prabhu2025vattention} decouples the allocation of virtual and physical memory using the CUDA virtual memory management approach and mitigates fragmentation while hiding the latency cost of on demand memory allocation.}
LightLLM~\cite{lightllm} takes a more granular token-level memory management mechanism to further diminish memory usage.
However, the overheads of such fragmented memory managing mechanisms pose new challenges.
Especially for cases where other optimizations are employed to boost the batch size, these fine-grained memory management methods might offer only marginal throughput benefits while substantially amplifying the inference latency.
It's evident that memory reduction in LLM inference is intricately tied with other algorithmic innovations and system-level optimizations.
While some might work well for specific workloads, they might counteract one another, leading to a degraded overall performance.
Striking the right balance between memory efficiency and computational performance of LLM inference systems remains an open and pressing challenge in the field.

\revision{There are also some approaches~\cite{aminabadi2022deepspeed,sheng2023high,miao2023specinfer,alizadeh2023llm,guo2023sti} enabling offloading techniques to use larger but slower memory (e.g., CPU DRAM, SSD) to save model parameters or KV cache in addition to the limited device memory (e.g., GPU DRAM).
For instance, systems like InfiniGen~\cite{lee2024infinigen} and HCache~\cite{gao2024fast} dynamically manage and offload KV cache to external memory tiers, thereby reducing GPU memory pressure and accelerating state restoration. Similarly, architectures such as Pensieve~\cite{yu2023stateful} and Mooncake~\cite{qin2024mooncake} disaggregate the inference pipeline to enable efficient migration and restoration of model states, while methods like CachedAttention~\cite{gao2024cost} exploit offloading to reuse attention computations across multi-turn conversations, striking a balance between performance and cost-efficiency in LLM serving.}

\subsubsection{Request Scheduling} 
\label{sec:sche}
Efficiently scheduling incoming inference requests is crucial for optimizing LLM serving. 
This section reviews request scheduling algorithms that maximize resource utilization, guarantee response time within latency service level objective (SLO), and handle varying request loads effectively.
Request scheduling for LLM serving shares commonalities with general ML serving techniques, as both aim to efficiently manage incoming requests and optimize resource utilization. These common aspects include dynamic batching~\cite{ali2020batch}, preemption~\cite{han2022microsecond}, priority~\cite{ng2023paella}, \revision{fairness~\cite{sheng2024fairness},} swapping~\cite{bai2020pipeswitch}, model selection~\cite{gunasekaran2022cocktail}, cost efficiency~\cite{zhang2019mark}, load balancing and resource allocation~\cite{weng2022mlaas}.
However, LLM serving also introduces unique challenges due to its distinctive characteristics, such as the massive model size, iterative autoregressive decoding mechanism, unknown variable output length and state management for context information.

Early LLM serving systems (e.g., FasterTransformer over NVIDIA Triton) only support request-level scheduling which is similar to prior approaches.
Orca~\cite{yu2022orca} first notices the gap between generative LLMs and the request-level scheduling of previous ML inference systems.
Considering the variable output sequence length, it schedules the execution of the engine at the granularity of iteration with a first-come-first-serve (FCFS) order and enables batching a selected set of operations for better hardware utilization.
Plenty of following approaches inherit the \textit{selective-batching} and \textit{iteration-level scheduling} policy, such as \textit{continuous batching} in vLLM and RayLLM~\cite{rayllm} and \textit{in-flight batching} in TensorRT-LLM~\cite{tensorrtllm}.
Moreover, SpecInfer extends to speculative decoding by iteratively selecting a batch of requests to perform one iteration of speculative inference and verification.
FastServe~\cite{wu2023fast} concentrates on the job completion time (JCT) and involves iteration-level preemption to prioritize requests with shorter input length, instead of FCFS.
Sarathi-Serve~\cite{agrawal2024taming} targets the pipeline bubbles in distributed inference caused by the initial iteration of varying length input requests. To saturate the GPU compute, it splits the input prompts into uniform chunks and piggybacks the chunk slot with other requests' decoding iterations if possible, which is also adopted by DeepSpeed-FastGen called Dynamic SplitFuse~\cite{deepspeedfastgen}.
S$^3$~\cite{jin2023s} involves an output sequence length predictor and helps to schedule more concurrent requests within the GPU memory constraint for larger batch size and higher inference throughput.
\revision{The prefill-decode disaggregation scheme was first proposed by Splitwise~\cite{patel2024splitwise} to split prefill from decode into different GPUs, while some concurrent work (e.g., DistServe~\cite{zhong2024distserve}, ExeGPT~\cite{oh2024exegpt}) have proposed similar architectures. LLM microserving~\cite{jin2024system} introduces a request-level programmable router to support dynamic reconfiguration of multiple disaggregation orchestration strategies.}

\revision{Mixed LLM serving workload often makes request scheduling more complex. For example, TetriInfer~\cite{hu2024inference} identifies the interference issue when serving heterogeneous downstream workloads (e.g., conversation, summarization, writing) with different length distribution. To address this problem, it utilizes the chunked prefill technique, applies the disaggregated architecture, and incorporates a two-level scheduling algorithm augmented with a length prediction model. Andes~\cite{liu2024andes} considers users' diverse Quality-of-Experience metrics for LLM-based text streaming services and introduces a preemptive request scheduling approach. ConServe~\cite{qiao2024conserve} executes online and offline LLM inference tasks simultaneously and leverages a priority-based scheduler to improve the overall resource utilization. Llumnix~\cite{sun2024llumnix} reschedules the heterogeneous requests react to the unpredictable workload dynamics at runtime based on a load-balancing scheduling policy and implements on top of inference engines.}

\subsubsection{Kernel Optimization} In this subsection, we delve into kernel-level optimizations, which target the performance of specific operations within the language model inference pipeline. These optimizations leverage hardware-specific features and software techniques to accelerate critical computation kernels.

\begin{itemize}
    \item \textbf{Kernel fusion}: To reduce overheads from kernel launching and memory accessing, kernel fusion is widely adapted by previous DNN frameworks and compilers.
    Since the backward computation is not required for LLM inference, more kernel fusion chances exist. Several contemporary Transformer inference engines (e.g., FasterTransformer~\cite{fastertransformer}, TenTrans~\cite{wu2021tentrans}, TurboTransformers~\cite{fang2021turbotransformers}, LightSeq~\cite{wang2020lightseq}, ByteTransformer~\cite{zhai2023bytetransformer}) and compilers (e.g. Welder ~\cite{shi2023welder}) propose to fuse 1) GEMMs with the same shape (e.g., the three linear transformations for query, key and value) and 2) Add Bias with the other non-GEMM kernels, such as residual connection, layer normalization and activation functions (e.g., ReLU).
    Among these, the optimization of fused multi-head attention kernel has been extensively explored and will be discussed in the following aspect.
    
    \item \textbf{Tailored attention}: 
    To make the attention operations run efficiently on a GPU, customizing or tailoring the GPU kernels specifically for the attention calculation is crucial.
    For example, cuDNN has provided a fused multi-head attention kernel API~\cite{cudnnmha}.
    Meanwhile, several implementations have been open-sourced for more performance gains.
    These can be roughly classified into two categories due to the special autoregressive decoding mechanism.
    One is for the first iteration (i.e., the initial/prefill/context/prompt phase), which processes all tokens from the input prompt in parallel. For example, xFormers~\cite{xFormers2022} extends the online softmax trick~\cite{rabe2021self,milakov2018online,choi2022accelerating} to the whole attention calculation using CUTLASS~\cite{cutlass}.
    The other is for the following iterations (i.e., the incremental/decode/generation phase) and the kernel only generates one output token per iteration.
    For autoregressive decoding, a common practice is to save the previously computed keys and values so that only a single query is required to compute when generating a new token instead of rerunning the entire sequence.
    The main direction of optimizations in this field is maximizing thread occupancy and minimizing the on-device high-bandwidth memory (HBM) access (i.e., using shared memory or registers~\cite{chen2021re}).
    They usually parallelize across the batch size and number of heads dimension (e.g., FasterTransformer) to distribute workloads.
    Some further enable parallelizing the sequence length dimension by partitioning the KV cache into chunks but require reducing the chunk-wise results at last, such as FlashDecoding~\cite{tri2023flash}.
    A subsequent work FlashDecoding++~\cite{hong2023flashdecoding++} removes such synchronization for partial softmax by introducing a unified maximum value known in advance.
    It is necessary to select the appropriate parallel dimension based on the workloads for better thread utilization.
    \revision{FlashInfer~\cite{ye2025flashinfer} utilizes the block-sparse format to unify diverse KV-Cache patterns, provides a customizable attention template to support different attention variants, and leverages CUDAGraph to maximize GPU utilization.}

    \item \textbf{Variable sequence length}: 
    Another unique challenge of LLM inference is that the sequences can vary in both input length and output length, and the latter is unknown in advance.
    One way to speed up inference is to process multiple sequences in a batch at once (\S\ref{sec:sche}). However, when a batch of sequences has variable input lengths, padding is often used to make them all the same length for batch processing, wasting computational and memory resources.
    To alleviate some of these inefficiencies, various strategies can be employed.
    Packing technique~\cite{packing,zhai2023bytetransformer} stores the sequences into a continuous memory space without padding and only unpacks before attention calculation.
    Ragged tensor~\cite{fegade2022cora} further supports computation with minimal padding using compiler-generated kernels.
    Bucketing the sequence into a smaller computation granularity (e.g., chunks~\cite{du2023improving}) is also a possible solution to alleviate memory usage of padding tokens. Due to the mixed execution of the initial phase and incremental phase, bucketing input prompts~\cite{agrawal2024taming} also brings new challenges to the memory management and request scheduling (\S~\ref{sec:sche}).

    \item \textbf{Automatic compilation}: Most existing LLM inference systems utilize vendor-specific libraries as their backend, such as cuBLAS, cuDNN and CUTLASS, which provide optimized kernel implementations. To further improve the inference efficiency, they also take great efforts on optimizing manually-written kernels for specific LLM operators (e.g., attention) over NVIDIA GPUs.
    Despite of these work, the trend of using automated DNN compilers still exists, such as TVM (i.e., Unity~\cite{sampson2022apache}, Relax~\cite{lai2023relax} and TensorIR~\cite{feng2023tensorir,ye2023sparsetir}), MLIR~\cite{katel2022mlir}, JAX~\cite{frostig2018compiling}, OpenAI Triton~\cite{tillet2019triton}, TASO~\cite{jia2019taso}, Mirage~\cite{wu2024mirage}, and TorchInductor~\cite{wu2023pytorch}. The compilation approach can help discover potentially more efficient operator implementations (e.g., expression derivation~\cite{zheng2023einnet}), and more importantly, facilitate adaptation to alternative hardware platforms, including mobile and edge devices, CPUs, DL accelerators, and other types of GPUs (e.g., AMD GPUs and Apple M2 Ultra).
    
\end{itemize}

\revision{In summary, our taxonomy categorizes the advancements in efficient LLM serving into two key dimensions: algorithmic innovations and system optimizations. Both dimensions play a critical role in reducing inference latency and enhancing throughput, though they approach these goals from different angles. The algorithmic side techniques streamline the generation process to accelerate inference while carefully managing the balance between speed and accuracy. Similarly, system-level optimizations refine the underlying computational framework to optimize resource utilization and further boost performance. When deploying these techniques in real-world applications, trade-offs between efficiency and accuracy must be carefully considered.
For example, in real-time conversational systems may  tolerate slight accuracy losses by using low-bit quantization for faster response times, whereas high-stakes applications like medical diagnostics demand the precision afforded by more computationally intensive methods.}

\section{Software Frameworks}
\label{sec:comp}

\begin{table}[]
    \caption{Comparison of state-of-the-art open-sourced GPU-based LLM serving systems. 
    }
    \label{tab:systems}
    \resizebox{\textwidth}{!}{
	\begin{tabular}{|l||c|c|c||c||c|c||c|c||c|c|p{5cm}|}
		\hline  
		\thead{Name\\Github} &  \multicolumn{3}{c||}{\thead{Parallel \\ Computation}} & \thead{Itera-\\tion-} & \multicolumn{2}{c||}{\thead{Attention \\ Kernel}} & \multicolumn{2}{c||}{\thead{Prioritized \\ Opt. Target}}  &  \\
	\thead{Ref.} & \thead{TP}  & \thead{PP}  & \thead{\tiny{Offload}} & \thead{Sche.} & \thead{Initial}  & \thead{Incremental}  & \thead{$L_{at}$}  & \thead{$T_{pt}$}   & \thead{Main Features} \\
		\hline 
        \makecell{FasterTrans-\\former~\cite{fastertransformer}} & $\surd$ & $\surd$ & &  & \makecell{\texttt{cuBLAS}\\\texttt{GEMM}} & \makecell{Fused\\attention} & $\surd$ & & \makecell[l]{
            \textbullet~ Manually-written kernel\\
            \textbullet~ Lightweight runtime
        }\\ \hline
        \makecell{FlexFlow-\\Serve~\cite{flexflowserve}}  & $\surd$ & $\surd$ & $\surd$ & $\surd$ & \makecell{\texttt{cuBLAS}\\\texttt{GEMM}} & \makecell{Tree\\attention} & $\surd$ & & \makecell[l]{
            \textbullet~ SpecInfer~\cite{miao2023specinfer}\\
            \textbullet~ Extremely low $L_{at}$
        }\\ \hline
        \makecell{vLLM~\cite{vllm}}  & $\surd$ & & $\surd$ & $\surd$ & \texttt{xFormers} & \makecell{Paged\\attention} & & $\surd$ & \makecell[l]{
            \textbullet~ Block-level KV cache~\cite{kwon2023efficient}\\
            \textbullet~ Enlarging batch size \& $T_{pt}$
        }\\ \hline
        \makecell{FlexGen~\cite{flexgen}}  &  & $\surd$ & $\surd$ &  & \makecell{\texttt{torch}.\\\texttt{bmm}} & \makecell{\texttt{torch}.\\\texttt{bmm}} & & $\surd$ & \makecell[l]{
            \textbullet~ CPU\&Disk Offload~\cite{sheng2023high}\\
            \textbullet~ Maximizing single GPU $T_{pt}$ 
        }\\ \hline
        \makecell{TGI~\cite{tgi}}  & $\surd$ & & & $\surd$ & \makecell{Flash\\attention} & \makecell{Paged\\attention} & & $\surd$ & \makecell[l]{
            \textbullet~ Huggingface integration
        }\\ \hline
        \makecell{DeepSpeed-\\Inference~\cite{dsinfer}}  & $\surd$ & $\surd$ & & & \makecell{\texttt{cuBLAS}\\\texttt{GEMM}} & \makecell{\texttt{cuBLAS}\\\texttt{GEMM}} & $\surd$ & & \makecell[l]{
            \textbullet~ Kernel auto-injection~\cite{dszeroinfer}\\
            \textbullet~ Multi-GPU \& Multi-Node
        }\\ \hline
        \makecell{ZeRO-\\Inference~\cite{dsinfer}}  & $\surd$ & $\surd$ & $\surd$ & & \makecell{\texttt{cuBLAS}\\\texttt{GEMM}} & \makecell{\texttt{cuBLAS}\\\texttt{GEMM}} & & $\surd$ & \makecell[l]{
            \textbullet~ CPU\&NVMe Offload~\cite{aminabadi2022deepspeed} \\
            \textbullet~ Maximizing single GPU $T_{pt}$
        }\\ \hline
        \makecell{Light-\\LLM~\cite{lightllm}}  & $\surd$ & & & $\surd$ & \makecell{Flash\\attention} & \makecell{Token\\attention} & & $\surd$ & \makecell[l]{
            \textbullet~ Token-level KV cache\\
            \textbullet~ Enlarging batch size \& $T_{pt}$
        }\\ \hline
        \makecell{MLC-\\LLM~\cite{mlc-llm}}  & $\surd$ & & &$\surd$ & \makecell{compiled\\MatMul} & \makecell{Paged\\attention} &  $\surd$ && \makecell[l]{
            \textbullet~ Universal deployment \\
            \textbullet~ Multiple types of GPUs
        }\\ \hline
        \makecell{TensorRT-\\LLM~\cite{tensorrtllm}}  & $\surd$ & $\surd$ & $\surd$ & $\surd$ & \makecell{\texttt{cuBLAS/}\\Flash-attn} & \makecell{Paged\\attention} & $\surd$ & & \makecell[l]{
            \textbullet~ NVIDIA Triton integration\\
            \textbullet~ Rich features supported
        }\\
\hline 
		
	\end{tabular} 
    }
\end{table}

Generative LLM serving requires a full stack of optimizations and many recent works have started to develop software frameworks to provide efficient LLM inference deployment service. In the following, we revisit these systems and investigate a comprehensive analysis of several representative open-sourced GPU-based LLM serving systems in Table~\ref{tab:systems}. The analysis does not contain some popular related projects, including 1) specialized solutions for other hardware (e.g., PopTransformer~\cite{poptrans}, CTranslate2~\cite{ctrans}, lammap.cpp and ggml~\cite{ggml}) and 2) deployment solutions built on top of the other systems, like OpenLLM~\cite{openllm} (vLLM), Xinference~\cite{xinference} (ggml + vLLM + xFormers), LMDeploy~\cite{lmdeploy} (FasterTransformer), gpt-fast~\cite{pytorchgptfast} (PyTorch), DeepSpeed-MII and DeepSpeed-FastGen~\cite{deepspeedmii} (DeepSpeed-Inference), and RayLLM and RayServe~\cite{rayllm} (vLLM).

We compare these state-of-the-art LLM serving systems and summarize their differences in several aspects.
First, most of these systems support tensor parallelism to enable multi-GPU inference and improve the system performance. And some of them future support pipeline parallelism or offloading to support inference over multi-node or resource-constrained environments individually. 
Second, partial systems learn from Orca and implement the iteration-level scheduling.
Third, we investigate the attention kernels of these systems and introduce their implementations in terms of the initial and incremental phases respectively.
For the initial phase, they usually adapt a batched general matrix multiply (GEMM) approach (e.g., cuBLAS, torch, Relay) and some utilize the online softmax trick to reduce HBM access (e.g., Flash-attention, xFormers). 
The incremental phase is more challenging because the per-token generation scheme results in lower computational intensity.
To improve the GPU utilization, FasterTransformer manually fuses the attention calculations (e.g., linear projection, positional bias, dot product, softmax, etc) into a single high-performance kernel template and involves several kernel optimization techniques, such as caching with shard memory, warp-shuffle instruction for reduction, half matrix multiplication and accumulation (HMMA) with tensor core and multiple-precision support.
FlexFlow-Serve enables speculative decoding and provides a tree-based parallel decoding kernel to verify the speculated tokens from multiple sequences (i.e., from multiple small models or different beams or parallel sampling) with zero-memory redundancy and maximum thread parallelism.
vLLM extends the fused mutli-head attention (MHA) kernel from from FasterTransformer by partitioning the KV cache into pages to eliminate redundant memory usage, especially for parallel sampling scenarios.
LightLLM takes a follow-up approach by partitioning the KV cache into more fine-grained token-wise pieces.

Note that, there still remain some other notable aspects that are not covered by the above discussions. For example, even for the most popular Flash and Paged attention kernels, they are usually implemented in different ways across these systems. TGI directly imports the original Flash/Paged attention libraries, LightLLM adopts kernels implemented by OpenAI Triton, MLC-LLM generates kernels by TVM, and TensorRT-LLM modifies from FasterTransformer's fused attention kernel to support paged attention.
Another example is about the input-aware kernel selection. For the initial phase, TensorRT-LLM selects from cuBLAS and Flash attention based on the context length. Besides the attention calculation, for the linear projection operators, there is also a recent trend of replacing GEMM with general matrix-vector product (GEMV) to handle the cases of small batch size (i.e., 1) more efficiently. And these systems also have many other different features, such as programming language (i.e., C++, Python), low-precision support (i.e., FP16, INT8), supported hardware and models. In summary, these different choices of design and implementation are largely determined by their prioritized optimization target. For example, vLLM proposes paged attention to improve the batch size for higher \textit{throughput} ($T_{pt}$), while FlexFlow-Serve leverages SpecInfer to accelerate decoding for lower \textit{latency} ($L_{at}$). 
Basically, low latency and high throughput are dual optimization targets in LLM serving systems, representing complementary but often conflicting objectives, necessitating a balanced strategy to optimize the trade-off between rapid response for individual tasks and maximizing the volume of tasks processed over a specified time frame.
Some recent studies~\cite{bestpractices} further decompose the response latency by TTFT+TPOT $\times$ output sequence length, where TTFT represents \textit{Time To First Token} and TPOT represents \textit{Time Per Output Token}. The former is driven by the initial phase processing speed while the latter directly depends on per-iteration execution time during incremental decoding.
Distinguishing these two metrics is beneficial to LLM service providers, leading to different system design choices and user experience (e.g., faster application responsiveness~\cite{liu2023cachegen}, longer prompts~\cite{deepspeedfastgen}).
\revision{Some recent kernel libraries and compilers (e.g., FlashInfer~\cite{ye2025flashinfer}, Mirage~\cite{wu2024mirage}, FlexAttention~\cite{dong2024flex}) provide more flexible program interfaces and generate optimized kernels adaptively according to the input configurations, which are also integrated with mainstream LLM serving engines.}
Although it unlikely to have a one-size-fits-all solution, we believe that future LLM serving systems will continually integrate these differentiated features, thereby continuously improving system efficiency and hardware utilization.

\section{Benchmarks}
\label{sec:bench}
\revision{Most existing work evaluate their system performance under real-world traces by leveraging the dynamic request arrival patterns from public production datasets (e.g., BurstGPT~\cite{wang2024burstgpt} and Azure~\cite{patel2024splitwise}.)}
Building a comprehensive and reproducible benchmark for comparing the performance of various LLM serving system like MLPerf~\cite{reddi2020mlperf} is a critical endeavor for both academic and industrial communities in this field.
It will not only help LLM users select the right system solutions but also encourage researchers and developers to keep pace with the advanced optimizations.
Unfortunately, despite of some prior reports~~\cite{hamel,llmperf}, up to this point, the community has not yet launched a convincing enough benchmark that takes into account all influencing factors. 
This is mainly because of the numerous evaluation settings, including model configuration, hardware environment, and request load, among others.
\revision{Most related benchmarks (e.g., LLM-Perf Leaderboard~\cite{llm-perf-leaderboard}) are designed for comparing the performance of different LLMs under the same infrastructure. LLM-Inference-Bench~\cite{chitty2024llm} has been proposed to evaluate the hardware inference performance of LLMs, including GPUs from NVIDIA and AMD and specialized AI accelerators, Intel Habana and SambaNova.}
Testing under a limited number of setting combinations cannot yield conclusions with credibility.
For example, certain system optimization techniques can only achieve performance advantages under high or low load conditions, and conversely, they might even be detrimental.
Besides, when measuring inference latency, how to exclude additional overheads not related to GPU inference (such as request scheduling overhead, inherent network latency, etc.) due to differences in system design is also a challenging topic.
Additionally, a fair benchmark test needs to consider the strict alignment of model output content, which is often overlooked in many tests.

\revision{Furthermore, designing such benchmarks demands careful calibration of trade-offs among throughput, latency, and cost-efficiency across a spectrum of workload scenarios --- from bursty and unpredictable traffic to sustained high-load conditions. A truly comprehensive evaluation must incorporate varied request patterns, diverse hardware configurations, and detailed measurements that isolate core inference performance from extraneous system overheads. Moreover, it is essential to account for differences in resource allocation, scheduling strategies, and communication latencies, ensuring that any performance gains are attributable to the underlying optimizations rather than artifacts of the testing environment. Establishing such a standard benchmark will provide valuable guidance for both practitioners and researchers, facilitating transparent comparisons and driving further innovation in LLM serving systems.}

\section{Connection with other surveys}
\label{sec:rela}

Our survey on efficient generative LLM serving and inference complements and extends the scope of existing literature in the field, while maintaining a distinct focus. Among the related works, \cite{kim2023full} comes closest in subject matter exploring the design of more general Transformer models and domain-specific accelerators. However, our survey differentiates itself by focusing specifically on generative LLM serving, a nuanced area that has not been the central focus of other studies.
Moreover, some studies delve into experimental investigations of LLM inference efficiency on GPUs~\cite{narayanan2023cheaply,zhang2023dissecting} and novel accelerators~\cite{emani2023comprehensive}, offering valuable empirical insights that are directly relevant to our focus on serving efficiency.
Additionally, LLMCarbon~\cite{faiz2023llmcarbon} addresses an increasingly important aspect of LLM deployment – its environmental impact (e.g., carbon footprints). While our survey's primary focus is efficiency from a performance standpoint, the environmental lens provided by such studies is undeniably relevant and respected in our broader discussion.
Some surveys and benchmarks~\cite{jaiswal2023compressing} offer valuable insights into model compression~\cite{zhu2023survey,gupta2022compression,zhu2023survey,treviso2023efficient} and quantization~\cite{yao2023comprehensive,gholami2022survey}. These studies lay a groundwork that indirectly supports our exploration of related directions. 
Some studies~\cite{muhlgay2023generating,dalvi2023llmebench} provide essential context for understanding LLM effectiveness (e.g., accuracy, perplexity, factuality and so on), which is beyond the scope of this survey. 
Our survey also acknowledges the contributions of prior surveys~\cite{ben2019demystifying,mayer2020scalable} focusing on distributed training of large-scale DNN models, as they inform the backdrop against which LLM serving must be considered.
In essence, our survey situates itself amidst a diverse array of studies, drawing from and contributing to a more holistic understanding of LLM serving efficiency, including both algorithmic innovations and system optimizations. By integrating insights from these various areas, we aim to provide a nuanced and comprehensive overview of the latest advancements and challenges in the field.

\section{Future Direction}
\label{sec:future}
As we stand at the forefront of LLM advancements, it becomes increasingly important to not only understand the current state of these technologies but also to anticipate and shape their future trajectory. Particularly in the realm of generative LLM serving, there is a vast landscape of unexplored possibilities and emerging challenges. The rapid evolution of this field necessitates a forward-looking approach, where identifying potential avenues for innovation and improvement is crucial. This foresight not only prepares us to adapt to upcoming technological shifts but also guides the research community toward addressing the most pertinent and impactful areas. In this context, we outline several promising directions for future research and development, each offering the potential to significantly enhance the efficiency of serving generative LLMs.

\boldparagraph{Development and Enhancement of Hardware Accelerators}
Future progress in enhancing generative LLM serving efficiency could be significantly driven by the development and refinement of specialized hardware accelerators, complemented by a co-design approach that aligns hardware and software optimizations.
For instance, integrating memory closer to processing units or optimizing chip architectures to better align with the data flow of LLM algorithms can lead to substantial reductions in latency and energy consumption. This approach has been exemplified in recent GPU advancements, like NVIDIA's Hopper architecture~\cite{nvhopper}, which demonstrates improvements in HBM and SRAM capacity, memory bandwidth, computing units and bisection bandwidth, directly benefiting the processing of LLMs.
Continued innovation in this area could involve designing hardware that is inherently tuned to the computational patterns of generative LLMs, such as optimizing for the specific demands of attention mechanisms and tensor operations that are prevalent in these models, eventually influencing the design and implementation of LLM serving systems.

\boldparagraph{Efficient and Effective Decoding Algorithms}
The development of more efficient decoding algorithms could substantially improve serving efficiency.
Motivated by the demand for more resource-efficient ways to utilize the vast knowledge encapsulated within LLMs, future work could explore alternative approaches to the traditional auto-regressive methods and unlock the generation speed for real-time applications while maintaining the decoding quality.
One promising direction is \textit{generalized speculative inference} as it enables preserving the same generation quality. Specifically, the small speculative model can be generalized to any other forms of methods that can generate draft tokens more efficiently than LLMs, such as knowledge retriever and user-defined functions~\cite{miao2023specinfer,yang2023inference}.
For example, some subsequent works arose recently, replacing the draft model with early exiting~\cite{yang2023predictive,zhang2023draft,bae2023fast,hooper2023speed} or non-autoregressive decoding~\cite{ge2022lossless,fu2023lookahead}. 
\revision{Some further generalize speculative decoding to handle other types of workloads, such as retrieval-augmented generation~\cite{zhang2024accelerating}, long input sequences~\cite{yang2025longspec,liu2025speculative}, and diffusion models~\cite{christopher2024speculative,janglantern}.}
In summary, the development of efficient decoding algorithms like speculative decoding coupled with the underlying system optimizations represents a significant opportunity to enhance the serving efficiency of generative LLMs.

\boldparagraph{Long Context/Sequence Scenarios Optimization}
As the application of LLMs continues to expand into more sophisticated scenarios, the demand for processing longer contexts or sequences is steadily growing. 
Serving LLMs with long-sequence workloads requires resolving the challenges from both the algorithm and system sides.
In terms of LLMs, they often suffer from \textit{length generalization failure} when sequences get longer than what was observed during training~\cite{press2021train} even enabling relative positional encoding~\cite{chen2023extending} or after fine-tuning on longer corpora~\cite{bai2023longbench}.
Even for some models that claim to support ultra-long contexts, studies have found that they encounter a situation of ``loss in the middle''~\cite{liu2023lost}.
Current approaches attempt to alleviate such limitations by reducing the computational sequence length while preserving relevant information, such as retrieval augmentation~\cite{xu2023retrieval}, sequence compression~\cite{jiang2023longllmlingua} and caching~\cite{gim2023prompt}.
For the LLM serving systems, longer sequence brings critical challenges, including more memory consumption and access of KV cache and quadratic increasing computational complexity of self-attention.

\boldparagraph{Investigating Alternative Architectures}
Although Transformer models and self-attention mechanisms currently dominate the landscape of LLMs, exploring alternative architectures is a promising direction for future research. The field of DL has historically seen a constant alternation of dominant architectures, with each new paradigm shift bringing about significant advancements. Given this trend, it's important to consider other architectural approaches that could offer distinct advantages, especially for improved computational efficiency.
For instance, some recent studies explore \textit{attention-free} methods~\cite{bozic2023rethinking}, using pure MLP (Multi-Layer Perceptron) architectures to replace attention mechanisms.
\revision{These changes also bring new innovation opportunities of the underlying inference engine, such as KV cache management~\cite{pan2024marconi}.}
The evolution of DNN model architecture is not only a natural progression, but also a necessary exploration to uncover more efficient and effective ways of structuring LLMs.

\boldparagraph{Exploration of Deployment in Complex Environments}
As the application of LLMs expands, a crucial future direction involves exploring and optimizing their deployment across various complex environments. This exploration goes beyond traditional cloud-based deployments to include scenarios like edge computing, hybrid computing (combining cloud and edge computing), decentralized computing, and the utilization of more affordable resources like spot instances.
Each of these environments presents unique challenges and opportunities for LLM serving. 
For instance, edge computing allows for faster response times and reduced bandwidth usage by processing data closer to the source, but it also poses challenges in terms of limited computational resources and storage capacity. 
Hybrid computing~\cite{hybridai} offers a balanced approach but requires advanced management to distribute computational tasks efficiently.
Decentralized computing presents a promising avenue for crowdsourcing computational resources, but it also brings additional considerations regarding data privacy and security~\cite{zhang2023latticegen,lu2023bumblebee}.
LLM serving over preemptive resources~\cite{asplos24spotserve} can significantly reduce monetary costs but requires fault tolerance mechanisms to handle their inherent unpredictability and variability, ensuring consistent performance and system reliability.
Successfully navigating the challenges from these complex environments will be key for more robust, scalable, and efficient LLM applications.

\boldparagraph{Automatic Adaptation to Specific Requirements}
The diverse application-specific requirements create a wide range of innovative LLM serving optimization opportunities, \revision{such as parameter-efficient fine-tuning~\cite{zhou2022pets,sheng2023s,chen2023punica,miao2024flexllm,wu2024dlora}, reinforcement or online learning and knowledge updates~\cite{sheng2024hybridflow,mei2024realhf,zhong2024rlhfuse}, retrieval from external vector storage~\cite{borgeaud2022improving}, multi-round conversation~\cite{lin2024parrot}, reasoning and planning~\cite{fu2024efficiently}, multi-modal workloads, structured generation~\cite{dong2024xgrammar,zheng2023efficiently}, and chaining together different LLMs' capabilities~\cite{wu2023autogen} (e.g., multi-agent simulation~\cite{xie2024ai})}. These unique challenges also demand automatic and smooth integration of LLM serving techniques into existing IT infrastructures by extending the optimization space to the whole LLM lifetime, including data acquisition and processing~\cite{miao2024demystifying}, AutoML~\cite{tornede2023automl} and model management~\cite{nagrecha2023saturn}, resource allocations, and performance monitoring. 
\section{Conclusion}

Efficient LLM serving is a fundamental step towards democratizing access to advanced AI technologies. This survey aims to provide researchers, practitioners, and developers with a comprehensive understanding of the existing methodologies, enabling them to make informed decisions when deploying LLMs in real-world environments. By consolidating the latest research findings on algorithms and systems, this survey paper hopes to accelerate progress and foster innovation in the pursuit of highly efficient LLM serving solutions.

\bibliographystyle{ACM-Reference-Format}
\bibliography{reference-clean}

\appendix











\end{document}